\pdfoutput=1
\documentclass{article}
\usepackage{iclr2027_conference,times}
\usepackage{amsmath,amssymb,graphicx,booktabs,multirow,tabularx}
\usepackage{xcolor,tikz,seqsplit}
\usepackage{hyperref,url}
\usepackage{microtype}
\usepackage{xurl}
\hypersetup{hidelinks}
\usetikzlibrary{positioning}
\newif\ifanonymous
\anonymousfalse
\ifanonymous\else\iclrfinalcopy\fi
\title{SAFE-MR: Evidence Sufficiency Learning for Selective Multimodal Rumor Detection}
\author{Shiwen Ni\\
Artificial Intelligence Research Institute,\\
Shenzhen University of Advanced Technology\\
\texttt{nishiwen@suat-sz.edu.cn}}
\newcommand{\R}{\mathbb{R}}
\begin{document}
\maketitle
% Override the camera-ready header: this manuscript is NOT an accepted paper.
\ifanonymous\else\lhead{Preprint}\fi
\begin{abstract}
Multimodal rumor detectors increasingly rely on retrieved evidence, yet relevant evidence is not necessarily sufficient for verification. Missing provenance, duplicated reports, and unresolved contradictions can produce confident predictions without adequate support. We introduce SAFE-MR, a framework that separates claim veracity from evidence sufficiency. The method decomposes image--text posts into verifiable claims, constructs a relation-aware claim--evidence graph, and aggregates evidence using provenance and contextual compatibility. Separate veracity and sufficiency heads support selective prediction, while evidence interventions encourage stability under irrelevant additions and sensitivity to evidence removal. On NewsCLIPpings, VERITE, and XFacta, SAFE-MR achieves macro-F1 scores of 91.2\%, 75.8\%, and 85.2\%, respectively. Against the matched backbone with evidence, its macro-F1 gains are 2.2, 4.9, and 4.8 percentage points. On the diagnostic selection set, SAFE-MR reduces AURC from 0.105 for maximum-probability rejection to 0.075 and lowers error at 80\% coverage from 13.8\% to 8.5\%. Evidence-perturbation and ablation results support the role of sufficiency learning and intervention training in improving selective verification.
\end{abstract}

\section{Introduction}
Authentic images can be repurposed to support false descriptions of when, where, or why an event occurred. Such out-of-context misinformation is difficult to detect from visual appearance alone: a flood photograph may be genuine even when its location and date are false. A detector must determine whether the image supports the specific contextual claim, rather than merely whether the image and text concern similar topics. Throughout this paper, multimodal rumor detection refers to image--text veracity assessment; it does not imply that every unverified rumor is false.

External retrieval offers a way to recover missing context. Evidence can reveal an image's earlier publication or establish the identity of a depicted event. However, retrieved pages may repeat the claim under investigation, omit decisive details, or disagree about the event. The number of matching pages is especially misleading when several pages copy a single source. Consequently, evidence relevance, source independence, and evidence sufficiency must be distinguished.

Recent methods provide strong foundations for evidence-based detection. SNIFFER investigates explainable out-of-context detection with multimodal language models~\citep{sniffer}; E2LVLM reorganizes retrieved evidence for model-based verification~\citep{e2}; TRUST-VL studies general misinformation detection across distortion types~\citep{trust}. Meanwhile, simple multimodal similarity features remain competitive on established benchmarks~\citep{muse}. These observations motivate evaluation beyond conventional accuracy: a system can exploit retrieval similarity without establishing whether the decisive factual proposition is supported.

\begin{figure}[t]
\centering
\input{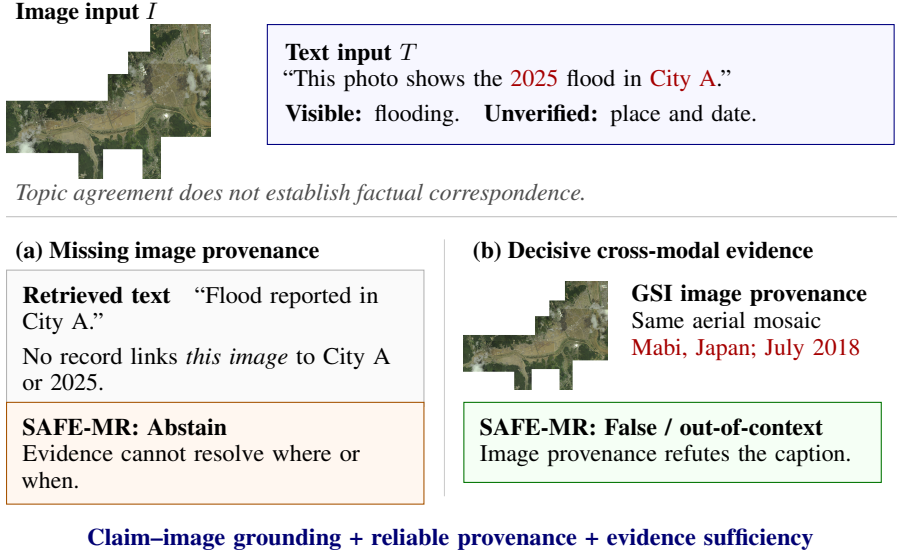}
\caption{Real imagery with a deliberately constructed incorrect caption. Left: topical evidence does not link the image to the claimed context. Right: the same mosaic's documented provenance (Mabi, Kurashiki, Japan; July 2018) refutes the attribution to 2025. Decisions illustrate intended behavior, not experimental outputs. Image credit: Geospatial Information Authority of Japan; mosaic assembled by Wikimedia Commons contributor Keikai; \href{https://creativecommons.org/licenses/by/4.0/}{CC BY 4.0}. The original mosaic is reproduced without cropping or content alteration. Source record: \href{https://commons.wikimedia.org/w/index.php?curid=70733654}{Wikimedia Commons}.}\label{fig:motivation}
\end{figure}

Consider the constructed mismatch in Figure~\ref{fig:motivation}, which uses a real aerial flood image. The deliberately incorrect caption attributes it to City A in 2025. A topical report alone does not establish the image's location or date. The source record instead identifies the mosaic as imagery of Mabi, Kurashiki, Japan, photographed by the Geospatial Information Authority of Japan on 9 July 2018 and subsequently assembled into a mosaic. This provenance refutes the proposed 2025 attribution. The claim's underlying truth has not changed; the evidence available to the verifier has changed. A conflicting repost does not invalidate decisive provenance evidence. This is a constructed illustration, not a documented social-media rumor or an evaluated model prediction.

We propose SAFE-MR to operationalize this distinction. It learns separate representations for veracity and evidence sufficiency, using atomic claims as the interface between image--text understanding and evidence reasoning. An intervention objective teaches the model which changes should preserve its output and which should reduce its willingness to decide. This is a controlled robustness objective, not a claim of identifying real-world causal effects.

Our contributions are threefold. First, we formulate evidence sufficiency as an evidence-conditioned target distinct from veracity and predictive confidence. Second, we develop relation-aware claim--evidence reasoning with provenance-sensitive aggregation and intervention-specific learning objectives. Third, we evaluate classification, evidence perturbations, selective prediction, and component ablations, showing how sufficiency modeling complements conventional confidence-based rejection.

\section{Related Work}
\subsection{Multimodal Consistency and External Verification}
NewsCLIPpings supports the study of contextually mismatched image--text pairs~\citep{news}. External-resource verification extends this setting by retrieving textual and visual evidence~\citep{retrieval}. However, similarity between a query and evidence is only a proxy for factual support. MUSE and AITR provide strong tests of whether architectural complexity contributes beyond this proxy~\citep{muse}. VERITE addresses unimodal biases through modality-balanced examples~\citep{verite}. We therefore evaluate both standard synthetic data and real-world transfer, with similarity-based and evidence-aware comparators.

\subsection{Vision--Language Models for Misinformation Detection}
SNIFFER~\citep{sniffer}, E2LVLM~\citep{e2}, and TRUST-VL~\citep{trust} motivate comparison with specialized vision--language models. Their explanations and evidence processing do not, by themselves, establish that confidence tracks the adequacy of the evidence supplied. XFacta additionally motivates separating retrieval quality from reasoning quality on recent real-world content~\citep{xfacta}. SAFE-MR focuses on evidence-conditioned sufficiency supervision and intervention-sensitive selection, complementing advances in backbone modeling and evidence retrieval.

\subsection{Evidence Quality and Selective Prediction}
Evidence pollution has already been investigated for out-of-context detection~\citep{pollution}, and FRAUD-Net considers variable or missing evidence~\citep{fraud}. These are necessary comparators for robustness claims. Selective prediction permits a classifier to abstain~\citep{selective}, while temperature scaling offers a straightforward confidence calibration baseline~\citep{calibration}. We compare SAFE-MR against these mechanisms applied to the same backbone. Improvements over a forced classifier alone would not establish that explicit sufficiency learning is useful.

\section{Materials and Methods}
\subsection{Task and Evidence-Conditioned Sufficiency}
Let $x=(I,T)$ be an image--text post, $E=\{e_k\}_{k=1}^{K}$ a retrieved evidence set, and $y\in\{0,1\}$ its veracity label, with 1 denoting false contextual attribution. We retain this binary truth label under evidence interventions. A separate target $a^*(x,E)\in\{0,1\}$ indicates whether the supplied, admissible evidence permits a justified decision. The action space is $\{0,1,\bot\}$, where $\bot$ denotes abstention and is not a third truth category.

Annotators judge sufficiency from the supplied evidence packet, not from unrestricted web access or remembered facts. Support requires coverage of the material claims and the image--event relation. Refutation may require only one decisive contradiction to a material claim. A source count or average coverage score is therefore insufficient as a complete decision rule. Unresolved conflict is recorded separately; disagreement from a demonstrably derivative or unreliable source need not force abstention.

\subsection{Claim Extraction and Evidence Preparation}
A vision--language backbone extracts a set $C=\{c_i\}_{i=1}^{m}$ of atomic propositions grounded in the post. Each proposition records a predicate, relevant entities, temporal and spatial qualifiers, and whether it is asserted by the text or inferred from visual content. Non-visible dates and locations are not treated as directly observed image facts. OCR is retained as a noisy observation with provenance. Each proposition also receives a materiality flag indicating whether contradicting it changes the post's contextual interpretation.

Textual queries combine material entities and event qualifiers, whereas image queries seek the image's publication context. Evidence records comprise textual content, optional images, and available provenance metadata. Near-identical or syndicated reports are treated as dependent rather than independent corroboration. Missing dates remain explicitly missing, and temporal compatibility concerns the event described in the claim rather than the recency of a retrieved page. The retrieval and grouping configuration is documented with the implementation record in Appendix~\ref{app:verification}.

\subsection{Relation-Aware Claim--Evidence Graph}
Claim and evidence encoders produce $h_i^c,h_k^e\in\R^d$. The graph connects candidate claim--evidence pairs selected by a high-recall retrieval stage. A learned relation predictor produces
\begin{equation}
 q_{ik}=\operatorname{softmax}\!\left(W_q[h_i^c;h_k^e;h_i^c\odot h_k^e;|h_i^c-h_k^e|]+b_q\right),
\end{equation}
over support, refute, and irrelevant. The unknown state is assigned at the aggregated claim level. Conflict is a relation between incompatible evidence statements about the same claim, rather than a mutually exclusive fourth stance of a single evidence item.

The reliability gate is
\begin{equation}
 r_k=\sigma\bigl(f_r(h_k^e,v_k)\bigr),
\end{equation}
where $v_k$ contains provenance, independence, temporal compatibility, and missingness features. Domain identity is an optional feature rather than a fixed truth oracle. Agreement is computed across provenance groups, so copied claims do not become independent corroboration.

For relation $r$, let $b_{ik}^{r}$ denote a claim--evidence compatibility logit and $W_r$ a relation-specific projection. Aggregation uses
\begin{equation}
 \alpha_{ik}^{r}=\frac{\exp(b_{ik}^{r})r_kq_{ik}^{r}}{\epsilon+\sum_{k'}\exp(b_{ik'}^{r})r_{k'}q_{ik'}^{r}},\qquad
 g_i=h_i^c+\sum_r\sum_k\alpha_{ik}^{r}W_rh_k^e.
\end{equation}
The normalized gate controls relative evidence contributions; by itself, it does not imply that uniformly reducing all reliability scores proportionally suppresses the aggregated representation. An explicit empty-evidence embedding is used when no evidence exists; the residual claim representation is retained. The model produces claim-level support, refute, and unknown probabilities from $g_i$. A separate conflict summary captures incompatible, credible statements that remain unresolved. These quantities are inputs to the decision heads, not hard-coded truth rules.

\subsection{Veracity, Sufficiency, and Selective Decision}
Materiality-aware pooling yields $g=\operatorname{Pool}(\{g_i\})$. We concatenate $g$ with claim coverage, decisive-refutation, conflict, and provenance-independence summaries into $z$. Two heads predict
\begin{equation}
 p_y=\operatorname{softmax}(W_yz+b_y),\qquad s=\sigma(W_sz+b_s).
\end{equation}
Here $s$ estimates $P(a^*=1\mid x,E)$. It is not defined as the number of retrieved pages or as maximum class probability. At inference, a positive calibration temperature $T_{\rm cal}$ defines $\widetilde p_y=\operatorname{softmax}((W_yz+b_y)/T_{\rm cal})$ and normalized uncertainty $u=H(\widetilde p_y)/\log 2$. The action is
\begin{equation}
 \hat d(x,E)=\begin{cases}\arg\max_y p_y(y),&s\geq\tau_s\ \text{and}\ u\leq\tau_u,\\\bot,&\text{otherwise}.\end{cases}
\end{equation}
Temperature is fitted on the held-out calibration partition; decision thresholds are selected on the separate tuning partition using temperature-calibrated uncertainty. Coverage-matched comparisons avoid rewarding a model merely for declining more examples. Calibration learned in-domain is not assumed to guarantee calibration after distribution shift.

\subsection{Evidence-Intervention Learning}
The base objective combines veracity, stance, and sufficiency supervision:
\begin{equation}
 \mathcal L=\mathcal L_{\rm ver}+\lambda_1\mathcal L_{\rm stance}+\lambda_2\mathcal L_{\rm suff}+\lambda_3\mathcal L_{\rm int}+\lambda_4\mathcal L_{\rm Brier}.
\end{equation}
Cross-entropy trains veracity and stance; binary cross-entropy trains sufficiency. Missing auxiliary labels are masked rather than imputed as negative. The Brier term is a proper scoring loss, not an ECE penalty. Temperature scaling is fitted after training.

Training interventions are constructed on training evidence; analogous perturbations are evaluated separately on held-out examples. Irrelevant addition preserves label and sufficiency when annotation confirms that no material information is introduced. We penalize prediction drift with Jensen--Shannon divergence and squared sufficiency differences. Key deletion lowers the sufficiency target only when all redundant decisive evidence is removed and the remaining packet becomes insufficient. Its ranking loss is $\max(0,\delta-s(E)+s(E^{-}))$, where $\delta=0.2$ is the ranking margin.

Conflicting addition receives packet-specific sufficiency labels: a weak repost opposing an authenticated archive may leave sufficiency unchanged, while two unresolved credible accounts can lower it. Pollution adds one group containing misleading evidence to a locally controlled packet, with the document's status recorded. These interventions do not require the veracity prediction to flip toward false evidence. The truth label remains fixed, while acceptance should decrease when justification is lost. Joint invariance of $(y,s)$ after decisive deletion would contradict the intended objective and is not imposed.

\begin{figure}[t]
\centering\resizebox{\linewidth}{!}{\begin{tikzpicture}[font=\fontsize{7}{8}\selectfont,>=stealth,
 ar/.style={->,black!60,semithick},c/.style={circle,draw=blue!60!black,fill=blue!12,inner sep=3pt},e/.style={draw=violet!60!black,fill=violet!12,inner sep=3pt}]
\newcommand{\doc}[3]{\draw[fill=#3!8,draw=#3!65] (#1,#2) rectangle +(.45,.6);\foreach \j in {1,2,3}{\draw[#3!60] (#1+.07,#2+.13*\j)--+(.3,0);}}
\newcommand{\mat}[3]{\foreach \i in {0,1,2}{\foreach \j in {0,1,2}{\filldraw[fill=#3!30,draw=white] (#1+.14*\i,#2+.14*\j) rectangle +(.14,.14);}}}
\node[anchor=west,font=\bfseries] at (0,1.5) {(a) Multimodal evidence reasoning};
\draw[fill=blue!5,draw=gray] (0,0) rectangle (1.1,.6);
\fill[gray!50] (.1,.1) rectangle (.3,.5);\fill[gray!35] (.75,.1) rectangle (.95,.53);
\fill[blue!35] (.01,.01) rectangle (1.09,.22);
\draw[fill=white] (.42,.17)--(.49,.35)--(.65,.35)--(.74,.17)--cycle;
\node at (.55,-.23) {Image $I$};
\foreach \i in {0,1,2,3}{\filldraw[fill=blue!15,draw=blue!60] (.12+.22*\i,-.9) rectangle +(.16,.22);}
\node at (.55,-1.15) {Text $T$};
\doc{.1}{-2.4}{violet}\doc{.55}{-2.25}{violet}\node at (.55,-2.65) {Evidence $E$};
\draw[draw=blue!60,fill=blue!6] (1.6,-1.1)--(2.1,-1)--(2.1,.55)--(1.6,.45)--cycle;
\node[rotate=90,font=\fontsize{5}{6}\selectfont] at (1.85,-.25) {Claim encoder};
\draw[draw=violet!60,fill=violet!6] (1.6,-2.5)--(2.1,-2.4)--(2.1,-1.65)--(1.6,-1.75)--cycle;
\node[rotate=90,font=\fontsize{5}{6}\selectfont] at (1.85,-2.08) {$f_e$};
\draw[ar] (1.12,.3)--(1.6,.3);\draw[ar] (1.05,-.8)--(1.6,-.8);\draw[ar] (1.05,-2.1)--(1.6,-2.1);
\mat{2.5}{-.05}{blue}\mat{2.5}{-.65}{blue}\mat{2.5}{-2.3}{violet}
\draw[ar] (2.12,-.25)--(2.45,-.25);\draw[ar] (2.12,-2.1)--(2.45,-2.1);
\node[align=center] at (4.2,.8) {Claim--evidence\\graph};
\node[font=\fontsize{5}{6}\selectfont,text=blue!60!black] at (2.72,.8) {$H_c$};
\node[font=\fontsize{5}{6}\selectfont,text=violet!60!black] at (2.72,-2.55) {$H_e$};
\node[font=\fontsize{5}{6}\selectfont,align=center,text=black!65] at (.57,-1.53) {who / where / when};
\draw[ar,dashed] (-.22,-.93)--(-.36,-.93)--(-.36,-2.05)--(.02,-2.05);
\foreach \i/\y in {1/0,2/-.8,3/-1.6}{\node[c] (c\i) at (3.55,\y) {$c_\i$};\node[e] (e\i) at (4.95,\y) {$e_\i$};}
\draw[green!50!black,thick] (c1)--(e1);\draw[green!50!black,thick] (c3)--(e1);
\draw[red!65!black,thick] (c2)--(e2);\draw[gray,dotted,thick] (c3)--(e3);
\draw[red!65!black,dashed] (e1.east) to[bend left=40] (e2.east);
\draw[ar] (2.96,-.25)--(3.25,-.25);\draw[ar] (2.96,-2.1)--(3.25,-1.82);
\node[align=center,font=\fontsize{3.8}{4.5}\selectfont,text=black!70] at (4.25,-2.65) {Green: support / Red: refute\\Dotted: irrelevant / Dashed: conflict};
\node[align=center] at (6.15,.8) {Reliability\\weighting};
\node[font=\fontsize{3.8}{4.5}\selectfont,text=black!65,align=center] at (6.7,-2.6) {source / time\\independence};
\foreach \y/\w in {0/.8,-.8/.45,-1.6/.15}{\fill[gray!10] (5.7,\y-.08) rectangle +(.9,.16);\fill[violet!50] (5.7,\y-.08) rectangle +(\w,.16);}
\node at (6.15,-2.1) {$r_k$};\draw[ar] (5.28,-.8)--(5.62,-.8);
\node[circle,draw=gray,inner sep=3pt] (sum) at (7,-.8) {$\sum$};\draw[ar] (6.65,-.8)--(sum);
\node[draw=blue!60,fill=blue!5,rounded corners,minimum width=1.55cm,minimum height=.55cm] (v) at (8.45,0) {Veracity $p_y$};
\node[draw=orange!70!black,fill=orange!7,rounded corners,minimum width=1.55cm,minimum height=.55cm] (s) at (8.45,-1.6) {Sufficiency $s$};
\draw[ar] (sum.east)--(7.45,-.8)|-(v);\draw[ar] (7.45,-.8)|-(s);
\node[circle,draw=gray,inner sep=3pt] (gate) at (9.85,-.8) {$\land$};
\draw[ar] (v.east)-|(gate.north);\draw[ar] (s.east)-|(gate.south);
\node[anchor=west,text=green!45!black] (a) at (10.5,-.25) {True / False};
\node[anchor=west,text=orange!70!black] (b) at (10.5,-1.35) {Abstain};
\draw[ar] (gate.east)--(10.25,-.8)|-(a);\draw[ar] (10.25,-.8)|-(b);
\node[font=\fontsize{6}{7}\selectfont] at (9.35,-2.35) {$s\geq\tau_s$ and $u\leq\tau_u$};
\fill[blue!45] (8,.43) rectangle +(.12,.28);
\fill[blue!22] (8.18,.43) rectangle +(.12,.12);
\node[anchor=west,font=\fontsize{5}{6}\selectfont,text=black!65] at (8.4,.55) {calibrated $p_y$};
\draw[black!35] (7.95,-2.05)--(8.9,-2.05);
\fill[orange!65] (8.55,-2.05) circle (.045);
\node[font=\fontsize{4.5}{5}\selectfont] at (7.95,-2.18) {0};
\node[font=\fontsize{4.5}{5}\selectfont] at (8.9,-2.18) {1};
\doc{10.8}{-2.7}{green}
\node[anchor=west,align=left,font=\fontsize{5}{6}\selectfont] at (11.31,-2.4) {Rationale $R$\\+ citations};
\draw[gray!35] (0,-3)--(11.9,-3);
\node[anchor=west,font=\bfseries] at (0,-3.45) {(b) Evidence-intervention learning};
\foreach \x/\t in {0/Irrelevant addition,4/Key deletion,8/Conflict or pollution}{\node[anchor=west] at (\x,-3.95) {\t};\doc{\x}{-4.85}{violet}\doc{\x+.55}{-4.85}{violet}\draw[ar] (\x+1.12,-4.55)--(\x+1.55,-4.55);}
\doc{1.7}{-4.85}{violet}\doc{2.25}{-4.85}{violet}\doc{2.8}{-4.85}{gray}
\doc{5.7}{-4.85}{violet}\draw[gray,dashed] (6.25,-4.85) rectangle +(.45,.6);\draw[red!65!black] (6.25,-4.85)--(6.7,-4.25);
\doc{9.7}{-4.85}{violet}\doc{10.25}{-4.85}{red}
\node[font=\fontsize{6}{7}\selectfont] at (1.6,-5.25) {$p_y'\approx p_y,\ s'\approx s$};
\node[font=\fontsize{6}{7}\selectfont] at (5.55,-5.25) {$s'\downarrow$ if decisive evidence is lost};
\node[font=\fontsize{6}{7}\selectfont] at (9.85,-5.25) {Reassess $s'$; keep truth label};
\draw[gray!25] (0,-5.65)--(11.9,-5.65);
\node[font=\fontsize{6}{7}\selectfont,text=black!70] at (5.95,-5.98) {$\mathcal L_{\rm ver}+\lambda_1\mathcal L_{\rm stance}+\lambda_2\mathcal L_{\rm suff}+\lambda_3\mathcal L_{\rm int}+\lambda_4\mathcal L_{\rm Brier}$};
\end{tikzpicture}}
\caption{SAFE-MR architecture. (a) Claim and evidence representations form a relation-aware graph, followed by reliability-aware aggregation and separate veracity and sufficiency heads. Conflict links connect incompatible evidence statements. Selection combines sufficiency with calibrated predictive uncertainty. (b) Training applies intervention-specific constraints: irrelevant additions preserve predictions, decisive deletions can lower sufficiency, and conflict or pollution retains the underlying truth label.}\label{fig:architecture}
\end{figure}
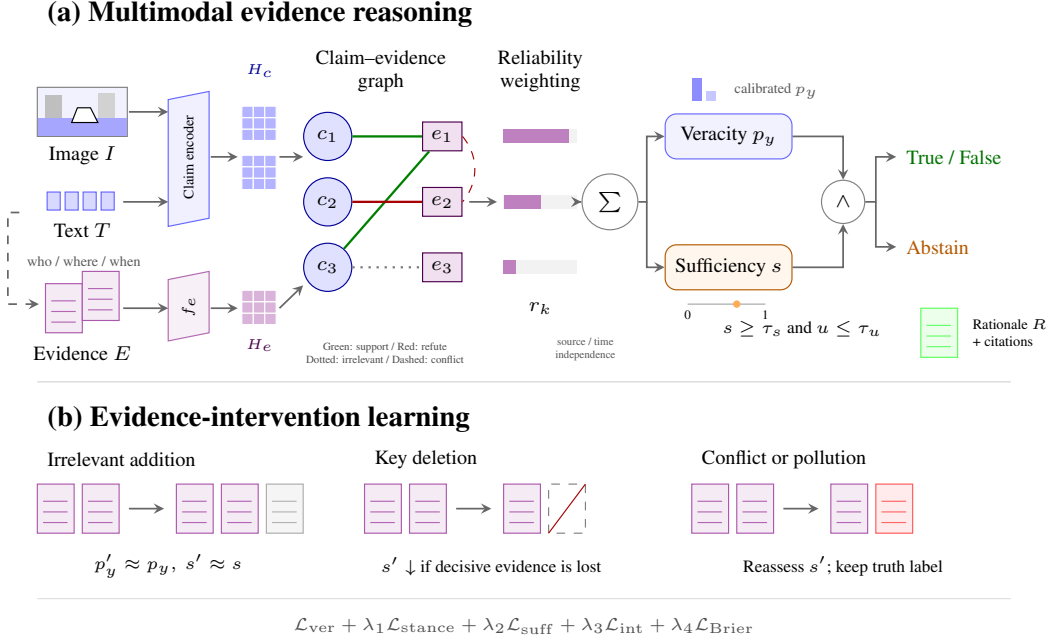

\subsection{Grounded Explanations}
The rationale component conditions on selected claims, evidence identifiers, predicted stances, and the decision. It can produce short statements with citations or describe missing evidence when the model abstains. This output is an interface for inspecting the decision, not a separately validated explanation metric: the quantitative evaluation below concerns classification and selection rather than semantic explanation faithfulness.

\section{Experiments}
\subsection{Datasets and Tasks}
We evaluate image--text verification on NewsCLIPpings~\citep{news}, VERITE-full~\citep{verite}, and XFacta~\citep{xfacta}. NewsCLIPpings provides synthetic out-of-context pairs. This study uses 100,000 training, 5,000 validation, and 5,000 test examples, with no sample exclusions. Each split is balanced between authentic matches and out-of-context mismatches (50\% each). These counts describe the split used in this study, rather than a universal official split. VERITE-full uses all 1,000 examples with binary labels: 338 truthful and 662 misleading examples, with both misleading categories mapped to false. The complete 1,000-example evaluation set is used throughout, without subset selection. VERITE posts do not enter model training, hyperparameter search, or temperature calibration. However, 1,200 evidence packets derived from all 1,000 posts were used before protocol freezing to develop annotation rules and clarify sufficiency judgments. Thus, VERITE is held out from model fitting but was not untouched during annotation-protocol development.

XFacta (arXiv:2508.09999v1) contains 2,400 examples, comprising an official development set of 240 examples (120 true and 120 false) and an official test set of 2,160 examples (1,080 true and 1,080 false). Its image manipulations, image--text mismatches, and misleading text make it a cross-type multimodal misinformation benchmark rather than an exclusively out-of-context task. Development and test memberships follow the official partition.

\begin{table}[t]\centering\small
\caption{Benchmark roles, evaluation populations, and task definitions. XFacta counts follow arXiv:2508.09999v1. Evidence-packet annotation partitions are separate from these benchmark partitions.}\label{tab:data}
\begin{tabularx}{\linewidth}{@{}lXX@{}}\toprule
Dataset & Population and split & Task\\\midrule
NewsCLIPpings & 100,000 train / 5,000 validation / 5,000 test; balanced classes; no exclusions & Binary image--text contextual verification\\
VERITE-full & All 1,000: 338 true / 662 misleading & Full-benchmark binary multimodal misinformation detection\\
XFacta & 2,400 total; development 240 (120/120); test 2,160 (1,080/1,080), true/false & Cross-type multimodal misinformation: image manipulation, image--text mismatch, and misleading text\\\bottomrule
\end{tabularx}\end{table}

\subsection{Sufficiency Annotation and Evidence Perturbations}
The model-development annotation set comprises 3,000 evidence packets constructed from the 240 official XFacta development posts and 860 additionally collected posts. Its 2,400 training, 300 tuning, and 300 calibration packets are partitioned by original post, so that all variants of a post remain in one partition. Separate resources comprise 1,200 VERITE-derived packets for annotation-rule development and 100 pilot packets for annotator training, totaling 4,300 packets outside the diagnostic suite. Pilot packets enter no formal split; VERITE-derived development packets enter no model-training, tuning, or calibration split.

GPT-5.6-Sol produces initial sufficiency labels, followed by individual human review. Two annotators perform first reviews, with each packet initially reviewed by one annotator. A random 600 packets (20\% of the 3,000-packet set) receive a second review by the other annotator, without access to the first human answer. Pre-adjudication agreement on binary sufficiency is 82\%. Of 108 disagreements, 36 are resolved through subsequent annotator review and 72 are adjudicated by the principal investigator.

These labels concern the adequacy of the supplied evidence for deciding a claim, not the underlying veracity label. Support requires evidence for the material contextual assertions; refutation may follow from a decisive contradiction. Repeated accounts from a common source do not constitute independent corroboration. An empty external packet is not automatically insufficient when the image itself resolves a directly visible assertion.

The diagnostic suite evaluates 600 base posts, with 200 from each benchmark, under nine conditions: clean evidence; evidence deletion at 25\%, 50\%, and 75\%; removal of all external evidence; irrelevant evidence addition; hard distractors; conflicting evidence; and evidence pollution. All 5,400 resulting packets are evaluated and manually labeled for sufficiency, stance, key evidence, and conflict. Diagnostic packets are distinct from the 4,300 packets above, and their posts are disjoint from model training, tuning, and calibration. The 200 VERITE diagnostic posts nevertheless overlap with posts used for early annotation-rule development, because that resource covered the full VERITE benchmark. Annotation rules and models are frozen before final evaluation.

Perturbations preserve the post and veracity label while changing its evidence. Table~\ref{tab:stress} evaluates 600 packets per condition. Tables~\ref{tab:selection} and~\ref{tab:ablation} pool all 5,400 packets, including clean cases. Variants of the same post are dependent observations; packet counts specify study scale rather than statistical adequacy.

\subsection{Comparison Methods}
The main classification comparison uses the matched backbone with evidence. It shares SAFE-MR's frozen CLIP ViT-L/14 backbone, evidence retrieval, and input data, but removes the graph module, sufficiency head, and intervention training, retaining only the veracity-classification head. Its parameter scale is matched to SAFE-MR. Exact parameter counts and the head configuration remain to be documented.

Table~\ref{tab:main} also reports local reproductions of CLIP feature fusion~\citep{clip}, MUSE and AITR~\citep{muse}, SNIFFER~\citep{sniffer}, FRAUD-Net~\citep{fraud}, DPOD~\citep{dpod}, E2LVLM~\citep{e2}, and TRUST-VL~\citep{trust}. Citations identify the methods; the reported numerical results are from this study rather than copied from the cited publications. All locally reproduced NewsCLIPpings baselines use the same 5,000-example test set. Evidence access is shared for the matched-backbone comparison but is not assumed identical across all other methods. The selector comparison includes maximum probability, temperature-scaled probability, a learned selector motivated by selective prediction~\citep{selective}, and SAFE-MR. Ablations remove sufficiency supervision, intervention training, reliability features, or atomic claims, or replace graph aggregation with flat fusion.

\subsection{Implementation and Metrics}
We use a frozen CLIP ViT-L/14 representation backbone and a two-layer graph module with hidden dimension 256 and dropout 0.1. Each example contains at most six atomic claims and five evidence groups. Claim extraction uses \texttt{Qwen/Qwen2-VL-7B-Instruct} with greedy decoding (\texttt{do\_sample=False}), an input limit of 4,096 tokens, and \texttt{max\_new\_tokens=512}. Explanation generation uses the same input limit and decoding policy with \texttt{max\_new\_tokens=256}.

The graph module and prediction heads are optimized with AdamW at a learning rate of $10^{-4}$ and weight decay 0.01. The effective batch size is 64, implemented as a single-device batch of 16 with four gradient-accumulation steps. Gradients are clipped at a global norm of 1.0. The graph module is trained in FP32; inference with the 7B vision--language model uses BF16 where supported. Training runs for at most ten epochs and stops early if the tuning loss does not improve for three consecutive epochs. The weights for veracity, stance, sufficiency, intervention, and Brier losses are 1, 0.5, 0.5, 0.3, and 0.1, respectively. The intervention ranking margin is $\delta=0.2$. The random seeds are 42, 43, and 44. For locally run experiments, metrics are computed separately within each seed on the corresponding evaluation population and then averaged across seeds; predictions from different seeds are not concatenated.

A single positive temperature is fitted by minimizing negative log-likelihood on the independent 300-packet calibration partition, separate from training and tuning. The fitted temperatures for seeds 42, 43, and 44 are 1.13, 1.08, and 1.17, respectively. A seed-42 pilot checkpoint, using the same temperature-calibration procedure, supplies predictions for joint threshold selection on the 300-packet tuning set. The grid spans $\tau_s\in[0.3,0.7]$ and $\tau_u\in[0.2,0.6]$, both in steps of 0.02. Selection maximizes binary sufficiency F1 for the joint decision $s\geq\tau_s$ and $u\leq\tau_u$ against the sufficiency target $a^*$. The selected thresholds, $\tau_s=0.52$ and $\tau_u=0.38$, are fixed for all three formal runs before test evaluation. Pilot threshold search, final rejection, and ranking all compute $u$ from temperature-calibrated classification probabilities.

Full-coverage evaluation reports accuracy and macro-F1 without removing rejected examples. SAFE-MR ranks examples by descending $q=s(1-u)$, with ties resolved by descending sufficiency probability $s$ and then ascending original sample identifier. At target coverage $c\in\{0.6,0.8\}$, the leading $\lfloor cn\rfloor$ examples are accepted. Coverage-matched risks and AURC use this ranking rather than the fixed double-threshold acceptance set.

Selective risk is the error fraction among accepted examples. Table~\ref{tab:selection} reports AURC, risk at 80\% and 60\% coverage, and two distinct ECE measures; AURC and ECE use the 0--1 scale, whereas risk is expressed as a percentage. For a ranked list with error indicators $e_{(i)}$, the discrete risk at coverage $k/n$ is $R_k=k^{-1}\sum_{i=1}^{k}e_{(i)}$. Discrete AURC is the arithmetic mean $n^{-1}\sum_{k=1}^{n}R_k$ over all nonempty prefix coverages. Reported AURC values remain those in the author's measured evaluation record.

ECE uses 15 equal-width bins. Classification-confidence ECE compares maximum class probability with the indicator of correct veracity prediction; the temperature-scaled baseline uses calibrated class probabilities. Sufficiency-probability ECE compares $s$ directly with the binary sufficiency truth $a^*$. Each selector's ECE is evaluated on its own accepted set; for SAFE-MR this is the set satisfying both fixed thresholds, not the top-60\% or top-80\% set. ECE values for different targets or different accepted populations do not constitute a controlled calibration comparison.

Sufficiency F1 and insufficient-evidence recall in Table~\ref{tab:ablation} use the manually annotated truth for the 5,400 diagnostic packets. The tables present point estimates. No unreported standard deviations, bootstrap intervals, multiplicity corrections, preregistration, or statistical significance tests are claimed.

\section{Experimental Results}\label{sec:results}
We evaluate full-coverage classification, evidence-perturbation robustness, selective prediction, and component contributions. Tables~\ref{tab:main}--\ref{tab:ablation} report the experimental results. Comparisons below describe the reported point estimates; statistical significance is not inferred from these values alone.

\subsection{Full-Coverage Classification}
\begin{table}[ht]\caption{Accuracy/macro-F1 (\%) at 100\% coverage for reproductions in this study. NewsCLIPpings uses the study's 100k/5k/5k train/validation/test split; all local baselines use the same test set. VERITE denotes the full 1,000-example binary mapping.}\label{tab:main}\small
\centering\renewcommand{\arraystretch}{1.12}
\begin{tabular*}{\linewidth}{@{\extracolsep{\fill}}lccc@{}}\toprule Method & NewsCLIPpings & VERITE-full & XFacta\\\midrule
CLIP feature fusion & 80.2/80.0 & 60.7/58.2 & 65.7/64.5\\
MUSE & 86.8/86.2 & 67.5/65.2 & 71.2/70.6\\
AITR & 88.2/88.0 & 70.3/68.5 & 75.3/74.7\\
SNIFFER & 87.4/87.2 & 70.0/68.8 & 74.0/73.5\\
FRAUD-Net & 88.5/88.1 & 71.2/69.7 & 76.0/75.3\\
DPOD & 88.2/88.0 & 70.3/68.8 & 75.3/74.6\\
E2LVLM & 89.4/89.0 & 73.0/71.9 & 81.2/80.5\\
TRUST-VL & 90.1/90.0 & 74.0/72.5 & 83.4/82.7\\
Matched backbone + evidence & 89.4/89.0 & 72.0/70.9 & 81.0/80.4\\
SAFE-MR & 91.5/91.2 & 77.4/75.8 & 86.4/85.2\\\bottomrule
\end{tabular*}

\end{table}

Table~\ref{tab:main} compares SAFE-MR with the matched backbone under shared representations, evidence retrieval, and inputs. SAFE-MR achieves macro-F1 scores of 91.2\%, 75.8\%, and 85.2\% on NewsCLIPpings, VERITE-full, and XFacta. Relative to the matched backbone, the gains are 2.2, 4.9, and 4.8 percentage points, respectively. These comparisons assess the combined contribution of the added reasoning and sufficiency components. VERITE results use the full 1,000-example binary mapping; its earlier use for annotation-rule development is disclosed in Section~4.1. The additional reproduced baselines are shown in the same table; citations identify their original methods, not the source of these reproduced values.

\subsection{Evidence Perturbations}
\begin{table}[ht]\caption{Pooled stress-test accuracy (\%) at full coverage; no rejected cases are removed. }\label{tab:stress}\small
\centering\renewcommand{\arraystretch}{1.12}
\begin{tabular*}{\linewidth}{@{\extracolsep{\fill}}lrrrr@{}}\toprule Condition & AITR & E2LVLM & Backbone & SAFE-MR\\\midrule
Clean & 77.4 & 81.2 & 80.5 & 84.6\\
Key deletion 25\% & 73.3 & 78.4 & 77.6 & 82.4\\
Key deletion 50\% & 69.2 & 74.6 & 73.5 & 80.3\\
Key deletion 75\% & 64.1 & 68.1 & 68.2 & 75.6\\
No external evidence & 60.4 & 62.7 & 63.5 & 64.3\\
Irrelevant addition & 72.6 & 77.5 & 75.7 & 82.8\\
Hard distractors & 66.7 & 71.3 & 69.4 & 78.5\\
Unresolved conflict & 65.5 & 70.7 & 68.3 & 76.4\\
Pollution & 66.4 & 72.4 & 69.8 & 78.2\\\bottomrule
\end{tabular*}\end{table}

Table~\ref{tab:stress} shows that SAFE-MR achieves the highest full-coverage accuracy in every reported condition. Its accuracy decreases from 84.6\% with clean evidence to 75.6\% under 75\% key-evidence deletion, a 9.0-point drop, compared with 13.1 points for E2LVLM and 12.3 for the matched backbone. SAFE-MR exceeds E2LVLM by 7.2 points under hard distractors and 5.7 under unresolved conflict. With no external evidence, its accuracy falls to 64.3\%, narrowing the margin over the backbone to 0.8 points. This pattern indicates that the gains depend substantially on the available evidence rather than uniformly stronger evidence-free predictions. Full-coverage robustness and selective reliability remain distinct properties; the latter is examined below.

\subsection{Selection and Calibration}
\begin{table}[ht]\caption{Selective prediction on the pooled diagnostic set. Risk is classification error (\%) at matched coverage; AURC and ECE are on the 0--1 scale. Class.\ ECE evaluates classification confidence; Suff.\ ECE evaluates sufficiency probability against $a^*$. ECE uses each selector's own accepted set, whereas risk uses the stated coverage. ECE columns and differing accepted populations must not be compared as a common calibration benchmark. Dashes denote unreported metrics.}\label{tab:selection}\small
\centering\renewcommand{\arraystretch}{1.12}
\begin{tabularx}{\linewidth}{@{}Xrrrrr@{}}\toprule Selector & AURC & Risk@80\% & Risk@60\% & \shortstack{Class.\\ECE} & \shortstack{Suff.\\ECE}\\\midrule
Maximum probability & 0.105 & 13.8 & 9.0 & 0.090 & --\\
Temperature + probability & 0.105 & 13.8 & 9.0 & 0.050 & --\\
Learned selector & 0.095 & 11.5 & 7.5 & -- & 0.055\\
SAFE-MR sufficiency + entropy & 0.075 & 8.5 & 5.5 & -- & 0.040\\\bottomrule
\end{tabularx}\end{table}

Table~\ref{tab:selection} separates selection quality from two calibration targets. Maximum-probability rejection and its temperature-scaled counterpart have identical AURC and coverage-matched risks of 0.105, 13.8\%, and 9.0\%. Their classification-confidence ECE values are 0.090 and 0.050, respectively, evaluated on their selector-specific accepted sets. SAFE-MR achieves an AURC of 0.075, a 28.6\% relative reduction from maximum-probability rejection and a 21.1\% reduction from the learned selector. At 80\% coverage, its error rate is 8.5\%, compared with 13.8\% for maximum probability and 11.5\% for the learned selector. At 60\% coverage, the corresponding rates are 5.5\%, 9.0\%, and 7.5\%. The learned selector and SAFE-MR have sufficiency-probability ECE values of 0.055 and 0.040 on their respective accepted sets. These assess calibration against $a^*$, not classification correctness, and do not establish calibration superiority across the four selectors. The AURC and coverage-matched risk improvements support evidence-conditioned selection beyond probability calibration alone.

\subsection{Sufficiency, Ablations, and Explanations}
\begin{table}[ht]\caption{Component ablations. F1 columns are percentages; AURC uses the 0--1 scale.}\label{tab:ablation}\small
\centering\renewcommand{\arraystretch}{1.12}
\begin{tabular*}{\linewidth}{@{\extracolsep{\fill}}lrrrr@{}}\toprule Variant & Stress F1 & Suff. F1 & Insuff. recall & AURC\\\midrule
Full SAFE-MR & 78.5 & 80.4 & 85.2 & 0.075\\
Without sufficiency supervision & 77.2 & 65.4 & 68.6 & 0.105\\
Without intervention training & 73.4 & 74.3 & 76.7 & 0.095\\
Without reliability features & 74.2 & 75.6 & 78.3 & 0.090\\
Without atomic claims & 75.1 & 76.4 & 79.6 & 0.088\\
Flat fusion, matched parameters & 76.3 & 77.6 & 80.5 & 0.085\\\bottomrule
\end{tabular*}\end{table}

The ablations in Table~\ref{tab:ablation} reveal complementary component contributions. Removing sufficiency supervision reduces stress macro-F1 by 1.3 points but increases AURC from 0.075 to 0.105, indicating a larger effect on selection than on forced classification. Removing intervention training yields the largest stress macro-F1 decrease, from 78.5\% to 73.4\%. Removing reliability features or atomic claims lowers stress macro-F1 by 4.3 or 3.4 points, respectively. Parameter-matched flat fusion reaches 76.3\% stress macro-F1 and 0.085 AURC, below the full model's 78.5\% and 0.075. The full model also achieves 80.4\% sufficiency F1 and 85.2\% insufficient-evidence recall. These auxiliary scores quantify agreement with the annotation targets and depend on the quality of the combined model and human labeling procedure.

The quantitative results in this section concern classification, sufficiency, selection, and calibration. They do not directly establish explanation faithfulness. Citation correctness, unsupported-assertion rates, and human judgments of explanation adequacy require separate evaluation and are not inferred from the reported tables.

\section{Discussion}
SAFE-MR is designed to expose an evidence boundary that a binary classifier usually hides. Its practical value would be a lower error rate at a fixed review workload, rather than a larger refusal rate in isolation. A verification service could route rejected examples to human reviewers with an explicit list of unresolved claims and missing provenance. Whether this improves reviewer efficiency requires a separate user study and is not established by offline metrics.

The design also raises a distinction between epistemic restraint and excessive conservatism. A reliable archive can refute a contextual caption even when some secondary details are unknown. Requiring every fact to be covered before any decision would unnecessarily reject such cases. Conversely, numerous low-quality repetitions should not provide sufficient support merely through agreement. These asymmetries motivate learned sufficiency labels and provenance grouping rather than a single average-coverage threshold.

The stress suite may overrepresent artificial evidence damage. For this reason, results must be paired with naturally incomplete packets and unseen events. A model trained on a particular pollution generator might recognize its style rather than learn verification. Generator-disjoint testing and paraphrase-preserving controls are needed before claiming robustness to synthetic evidence. Interventions remain local research fixtures; they are not published as apparently authentic news pages.

\section{Limitations}
The sufficiency target depends on judgments of claim materiality and evidence admissibility. Combined model and human annotation can retain systematic ambiguities or biases, particularly for conflicting accounts. Search-engine coverage and source-language availability can also disadvantage some claims. Domain reputation must not be equated with truth, and historical timestamps may be missing or incorrect.

Pretrained models may have encountered benchmark claims or fact-checking pages. A downstream test split does not remove this possibility, and the present results do not establish temporal generalization or absence of pretraining contamination. The evaluation concerns static image--text verification; video manipulation, propagation modeling, and multilingual multi-image reasoning are outside its primary scope.

Point estimates alone do not characterize run-to-run uncertainty. Synthetic evidence perturbations also approximate only part of the missingness and disagreement encountered in deployment. Explanation quality and the effect on human reviewer efficiency require separate evaluation.

\section{Conclusions}
We presented SAFE-MR, a selective multimodal verification framework that separates veracity prediction from evidence sufficiency. Atomic claims organize evidence, reliability-aware aggregation accounts for provenance, and intervention training distinguishes irrelevant changes from losses of informative evidence. Across the three classification benchmarks, SAFE-MR improves macro-F1 over the matched backbone with evidence. On the diagnostic set, it reduces AURC and coverage-matched error relative to confidence-based rejection and a learned selector. Ablations associate sufficiency learning primarily with better selection and intervention training with greater perturbation robustness. Together, these results support explicit evidence-sufficiency modeling as a useful complement to multimodal truth prediction.

\ifanonymous\else
\section*{Funding}
Shiwen Ni is supported by the Shenzhen Science and Technology Program (\seqsplit{JCYJ20250604182917023}), Shenzhen Major Science and Technology Project (\seqsplit{KCXFZ20240903094007010}), and GuangDong Basic and Applied Basic Research Foundation (2026A1515060001, 2024A1515012003 and 2023A1515110718).
\fi

\section*{Data and Model Use}
The benchmark datasets are distributed by their original providers. GPT-5.6-Sol was used alongside human annotation to construct evidence-sufficiency targets. AI-assisted drafting and figure preparation were used during manuscript preparation. Dataset redistribution remains subject to the original licenses. No ethics approval identifier, annotator consent arrangement, or code-release URL is asserted without an accompanying record.

% Compilation: place the official ICLR 2027 iclr2027_conference.sty
% in the same directory as manuscript.tex.
% For a BibTeX-based version: pdflatex manuscript.tex; bibtex manuscript;
% pdflatex manuscript.tex; pdflatex manuscript.tex; pdflatex manuscript.tex.
% This version uses inline thebibliography: skip BibTeX (no external .bib)
% and run pdflatex repeatedly.
\clearpage
\appendix
\section{Implementation and Evaluation Details}\label{app:verification}
\subsection{Dataset Partitions and Evaluation Units}
VERITE-full retains the complete official 1,000-example benchmark, with 338 truthful examples and 662 examples mapped from the two misleading categories to the false class. No VERITE subset is used. XFacta follows arXiv:2508.09999v1: 240 development examples (120 true and 120 false) and 2,160 official test examples (1,080 true and 1,080 false), totaling 2,400. Its evaluation spans image manipulation, image--text mismatch, and misleading text. The NewsCLIPpings split used in this study contains 100,000 training, 5,000 validation, and 5,000 test examples. All three splits contain equal proportions of authentic matches and contextual mismatches, with no sample exclusions.

The 1,200 VERITE rule-development packets cover all 1,000 original VERITE posts. Consequently, the 200 VERITE diagnostic posts share base posts with this early resource, although their evidence packets differ. These posts never enter model training, hyperparameter search, or temperature calibration. Their reuse for annotation-rule development means that VERITE is not an untouched test resource across the full research process. Rules and models are frozen before final evaluation. Benchmark membership, annotation membership, and diagnostic membership are recorded at both post and packet levels. Post identifiers define grouping across evidence packets and their variants. An exclusion record must identify the affected post, partition, and reason, such as an unavailable or unreadable required input; exclusions must not depend on prediction correctness. Empty external evidence in the designated no-evidence condition is a valid diagnostic input, not an exclusion. Any execution-specific exclusion must be reconciled with the evaluation denominator. No NewsCLIPpings samples are excluded.

\subsection{Annotation Pipeline}
GPT-5.6-Sol supplies initial evidence-sufficiency labels. Qwen2-VL-7B-Instruct is used for claim extraction and explanation generation, not initial sufficiency labeling. The formal 3,000-packet annotation set is constructed from 240 XFacta development posts and 860 additional posts, grouped by post into training (2,400 packets), tuning (300), and calibration (300). A separate 100-packet pilot enters no formal partition. The 1,200 VERITE-derived rule-development packets are separate from both resources. Together these comprise 4,300 packets outside the diagnostic suite.

Two human annotators review the formal annotations. A randomly selected 600-packet subset receives independent second review without the first human answer. Agreement before resolution is 82\%; 108 packets disagree, with 36 resolved by subsequent annotator review and 72 adjudicated by the principal investigator. Diagnostic sufficiency, stance, key-evidence, and conflict targets are manually annotated for all 5,400 diagnostic packets. No Cohen's kappa value is reported without a verified calculation from the original paired labels.

Sufficiency is judged relative to the supplied packet and material claim. Duplicate reports from one original source form one provenance group. Support requires coverage of material assertions, whereas decisive contradictory evidence can establish refutation. The packet counts specify the study scale and do not establish statistical sufficiency.

\subsection{Model and Optimization Configuration}
The representation backbone is frozen CLIP ViT-L/14. The graph module has two layers, hidden dimension 256, and dropout 0.1. Each example is capped at six atomic claims and five evidence groups. The claim-extraction checkpoint is \texttt{Qwen/Qwen2-VL-7B-Instruct}. Greedy decoding uses \texttt{do\_sample=False}, at most 4,096 input tokens, and output limits of 512 new tokens for claim extraction and 256 for explanations.

AdamW updates the graph module and prediction heads with learning rate $10^{-4}$, weight decay 0.01, and global gradient-norm clipping at 1.0. An effective batch size of 64 comprises a single-device batch of 16 and four accumulation steps. Graph training uses FP32; vision--language inference uses BF16 when supported. Training lasts at most ten epochs with early stopping after three consecutive epochs without tuning-loss improvement. Loss weights are 1 for veracity, 0.5 for stance, 0.5 for sufficiency, 0.3 for intervention, and 0.1 for Brier loss. The intervention margin is $\delta=0.2$. Seeds are 42, 43, and 44. Locally run results average the metrics computed separately for these three seeds. Pooled diagnostic metrics combine samples across datasets and conditions within a seed, not across seeds.

\subsection{Calibration, Ranking, and Perturbation Rules}
One positive temperature minimizes negative log-likelihood on the independent 300-packet calibration partition. Temperatures are 1.13, 1.08, and 1.17 for seeds 42, 43, and 44. The seed-42 pilot uses the same calibration procedure before joint threshold search on the separate 300-packet tuning set. The search grids are $\tau_s\in[0.3,0.7]$ and $\tau_u\in[0.2,0.6]$ with step 0.02. Binary sufficiency F1 of the joint threshold decision selects $\tau_s=0.52$ and $\tau_u=0.38$, fixed for all three formal runs.

Both final rejection and ranking use uncertainty from temperature-calibrated class probabilities. Ranking sorts $q=s(1-u)$ descending, then $s$ descending, then original sample ID ascending. At coverage $c$, the first $\lfloor cn\rfloor$ samples are accepted. AURC is the arithmetic mean of risks over all nonempty ranked prefixes.

Both ECE definitions use 15 equal-width bins but different targets. Classification-confidence ECE uses maximum class probability and classification correctness; sufficiency-probability ECE uses $s$ and binary sufficiency truth $a^*$. Each value is conditional on the corresponding selector's accepted set. SAFE-MR's ECE set satisfies $s\geq0.52$ and $u\leq0.38$. These ECE populations are separate from the fixed-coverage sets used for risk reporting.

The diagnostic design comprises 600 base posts, 200 per benchmark, evaluated in nine conditions to yield 5,400 packets. Clean packets retain the original evidence. Deletion operates on evidence items, removing $\lfloor rK\rfloor$ of the $K$ original items for $r\in\{0.25,0.50,0.75\}$; the no-external-evidence condition removes all external evidence while preserving the post. Irrelevant additions introduce evidence unrelated to the material claim. Hard distractors are superficially relevant but do not resolve the claim. Conflict introduces incompatible accounts; pollution introduces misleading external evidence. Each construction preserves the original image--text post and veracity label.

Irrelevant, hard-distractor, conflict, and pollution conditions each add one evidence group. After addition, if the packet contains more than five groups, the top five by relevance are retained. Thus, addition can displace an original group when the cap is reached. The particular deletion masks, distractor sources, and conflict/pollution content follow the perturbation records. Post and variant identifiers distinguish the 600 base observations from their 5,400 dependent packets. Any uncertainty analysis must preserve this post-level dependence. Table~\ref{tab:stress} evaluates 600 packets per condition. Tables~\ref{tab:selection} and~\ref{tab:ablation} use all 5,400 packets, including clean cases. Metrics are computed on the pooled examples within each seed and then averaged over seeds 42, 43, and 44.

\subsection{Measured Scores and Engineering Estimates}
Reported performance scores are experimental measurements. AITR achieves 88.2\% accuracy and 88.0\% macro-F1 on the NewsCLIPpings test split used in this study.

Hardware capacity and human workload below are engineering estimates, not experimental results or measured efficiency claims. The estimated resource configuration is one 24\,GB GPU for graph-module training and one 48\,GB GPU for Qwen2-VL-7B BF16 inference. Human first review is estimated at 2--5 minutes per packet, corresponding to approximately 100--250 person-hours for 3,000 packets. This estimate covers only first review of the 3,000 model-development packets, not the entire annotation effort. Repeated review, dispute adjudication, the 1,200 VERITE rule-development packets, the 100-packet pilot, and annotation of the 5,400 diagnostic packets require separate accounting.

\end{document}